\documentclass{article}

\usepackage[utf8]{inputenc}
\usepackage{amsmath,amssymb,mathtools}
\usepackage{booktabs,multirow,makecell,tabularx,array}
\usepackage{graphicx}
\usepackage{float}
\usepackage{algorithm}
\usepackage{algpseudocode}
\usepackage{xspace}
\usepackage{url}
\usepackage{microtype}
\usepackage{placeins}
\usepackage[final]{corl_2026}
\makeatletter
\renewcommand{\@conferencelocation}{Austin, Texas, USA}
\makeatother
\hypersetup{
    pdftitle={GeoAlign: Beyond Semantics with State-Guided Spatial Alignment in VLA Models},
    pdfauthor={Yizhi Chen, Zhanxiang Cao, Xinyi Peng, Yixiao Zheng, Xiaxi Si, Yiheng Li, Liyun Yan, Keqi Zhu, Xueyun Chen, Shengcheng Fu, Tianyue Zhan, Yufei Jia, Jinming Yao, Yan Xie, Kun Wang, Cewu Lu, Yue Gao}
}

\newcommand{\method}{GeoAlign\xspace}
\newcommand{\gep}{GEP\xspace}
\newcommand{\R}{\mathbb{R}}
\newcommand{\E}{\mathbb{E}}
\newcommand{\Dpol}{\mathcal{D}_{\mathrm{pol}}}
\newcommand{\Ddep}{\mathcal{D}_{\mathrm{dep}}}
\newcommand{\MHA}{\mathrm{MHA}}
\newcommand{\LN}{\mathrm{LN}}
\newcommand{\FFN}{\mathrm{FFN}}

\newcommand{\reshape}{\mathrm{reshape}}
\newcommand{\cmark}{\ensuremath{\checkmark}}
\newcommand{\xmark}{\ensuremath{\times}}
\newcommand{\pmark}{\ensuremath{\triangle}}
\newcommand{\res}[1]{#1}
\newcolumntype{Y}{>{\raggedright\arraybackslash}X}
\newcolumntype{L}[1]{>{\raggedright\arraybackslash}p{#1}}

\title{GeoAlign: Beyond Semantics with State-Guided Spatial Alignment in VLA Models}
\author{
{\small\bfseries
Yizhi Chen$^{1,2}$ \quad Zhanxiang Cao$^{3,2}$ \quad Xinyi Peng$^{1,2}$ \quad
Yixiao Zheng$^{7}$ \quad Xiaxi Si$^{3,2}$}\\
{\small\bfseries
Yiheng Li$^{3,2}$ \quad Liyun Yan$^{3,2}$ \quad Keqi Zhu$^{4,2}$ \quad
Xueyun Chen$^{5}$ \quad Shengcheng Fu$^{1,2}$}\\
{\small\bfseries
Tianyue Zhan$^{3,2}$ \quad Yufei Jia$^{6}$ \quad Jinming Yao$^{8}$ \quad
Yan Xie$^{7}$ \quad Kun Wang$^{7}$}\\
{\small\bfseries
Cewu Lu$^{3,2}$ \quad Yue Gao$^{3,2,*}$}\\[0.35em]
{\normalfont\small
$^{1}$Tongji University \quad
$^{2}$Shanghai Innovation Institute \quad
$^{3}$Shanghai Jiao Tong University}\\
{\normalfont\small
$^{4}$Zhejiang University \quad
$^{5}$Jingdezhen Ceramic University \quad
$^{6}$Tsinghua University}\\
{\normalfont\small
$^{7}$HONOR \quad
$^{8}$University of Science and Technology of China}\\[0.25em]
{\normalfont\small
Yizhi Chen: \href{mailto:15970600542@163.com}{\texttt{15970600542@163.com}}}\\
{\normalfont\small
$^*$Corresponding author: \href{mailto:yuegao@sjtu.edu.cn}{\texttt{yuegao@sjtu.edu.cn}}}
}
\begin{document}
\maketitle

\begin{figure}[H]
    \centering
    \includegraphics[width=\linewidth]{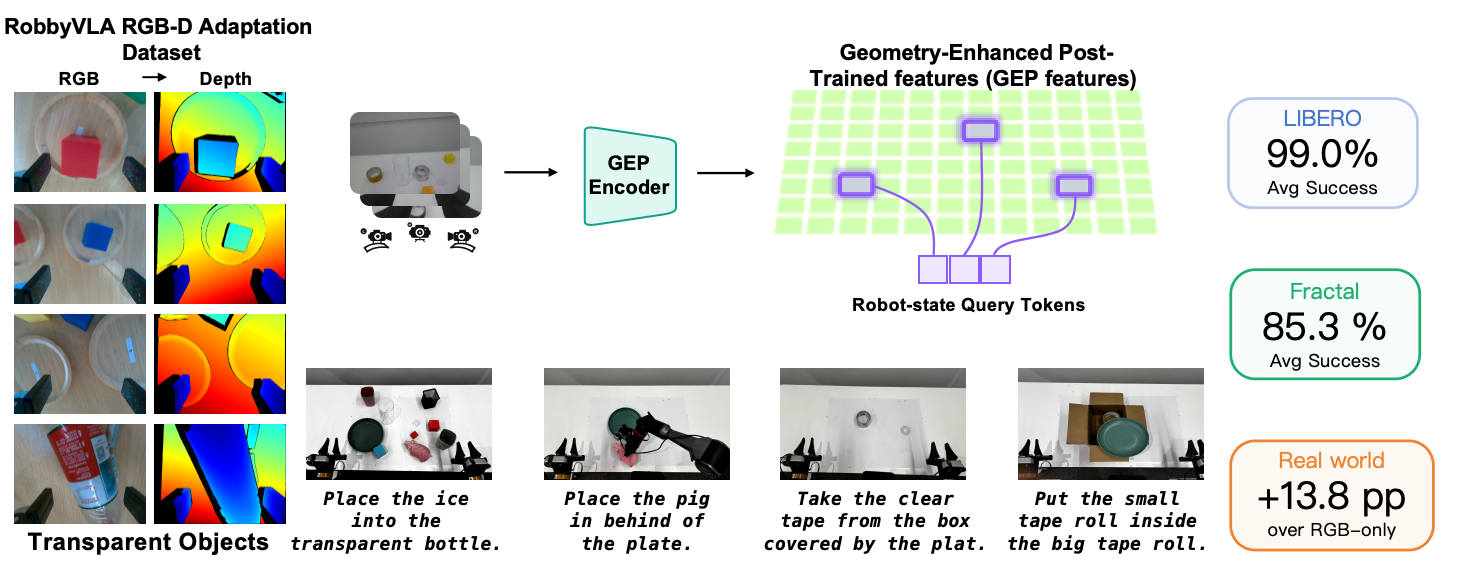}
    \caption{\textbf{GeoAlign at a glance.} \method learns RGB-derived geometry features and uses proprioceptive-state queries to extract compact geometry tokens for a VLA action decoder.}
    \label{fig:begin_teaser}
\end{figure}

%===============================================================================
\begin{abstract}
Current Vision--Language--Action (VLA) models often optimize for semantic grounding, whereas executable manipulation requires geometry-aware spatial alignment. We introduce GeoAlign, a state-guided spatial alignment architecture for VLA policy learning. Offline robot-domain RGB-D supervision post-trains an RGB geometry branch to produce Geometry-Enhanced Post-Trained (GEP) features; the depth head is then discarded, so policy training and rollout use RGB, language, and proprioceptive state. State-generated queries attend to the GEP grid to produce eight compact geometry tokens for action prediction. GeoAlign achieves \res{99.0}\% on LIBERO, \res{85.3}\% across three SimplerEnv-Fractal task families, and \res{78.8}\% on eight real-world ALOHA tasks, with ablations supporting the combined recipe of geometry post-training and state-guided querying on Isaac-GR00T N1.6-3B.
Project website: \url{https://chenyizhi123.github.io/geoalign-project/}.
\end{abstract}
\keywords{Learning representations for robotic perception and control, Robot manipulation, Vision--language--action}

%===============================================================================
\section{Introduction}
\label{sec:introduction}

A long-standing goal in robotics is to build generalist robot policies that follow language instructions, handle diverse objects and scenes, and adapt across tasks and embodiments. Vision--language--action (VLA) models have become a promising route toward this goal by adapting pretrained vision--language representations with robot demonstration data to map observations and instructions to actions~\citep{brohan2023rt1,zitkovich2023rt2,oneill2023openx,kim2025openvla,black2024pi0,black2025pi05,bjorck2025gr00tn1}. Recent work expands this paradigm through heterogeneous robot data, diverse action interfaces, and predictive or generative world-model representations for manipulation~\citep{black2024pi0,liu2024rdt,bjorck2025gr00tn1,zhang2025dreamvla,cen2025worldvla,chen2025goalvla,yang2026rise,tian2026starry}.

Despite this progress, fine-grained manipulation still depends on spatial details that are easy to lose in high-level semantic representations. Current VLA models often optimize for semantic grounding, whereas executable manipulation requires geometry-aware spatial alignment and dynamic affordance selection. A policy may identify the correct object and parse the instruction, yet fail when the action requires tight clearance, precise alignment, contact-sensitive motion, stable release, or geometry reasoning over transparent and annular objects. These failures reflect a local geometry problem in which the action decoder must be guided by spatial features that determine whether the next action chunk is physically executable~\citep{gervet2023act3d,zhang2024dynamiccompliance,zhao2025anyplace,nadeau2025stableplacement}.

Recent work has explored two complementary routes to address the gap between semantic grounding and executable manipulation. One route strengthens the perception side of VLA by enriching visual-language representations with stronger spatial understanding through depth-aware modules, 3D region context, spatial grounding, or auxiliary depth supervision~\citep{chen2024spatialvlm,song2025robospatial,yuan2025depthvla,li2025qdepthvla}. These methods improve the spatial content available before action decoding, making visual representations less purely semantic and more geometry-aware. However, methods that rely on measured depth or explicit 3D inputs can inherit sensing failures around transparent or thin-structured objects~\citep{sajjan2019cleargrasp}.

\begin{figure}[!t]
    \centering
    \includegraphics[width=\linewidth]{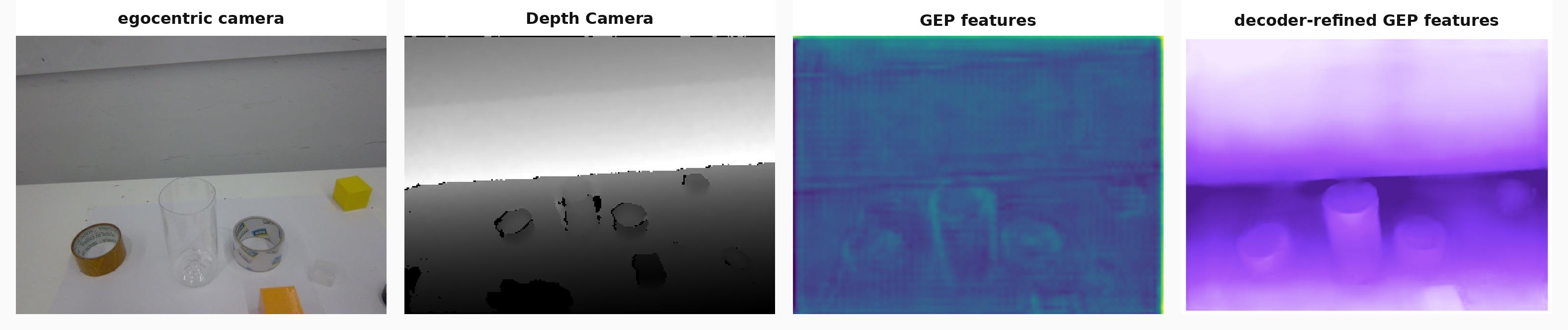}
    \caption{\textbf{RGB-derived geometry under degraded measured depth.} RGB observations preserve transparent and annular object structure, while the measured depth map contains missing or fragmented regions. \gep features retain image-space geometry cues used for policy conditioning.}
    \label{fig:gep_transparent}
\end{figure}

A complementary route incorporates spatial structure directly into policy execution and action generation. Prior manipulation and VLA methods have attempted to make action generation more spatially grounded through explicit spatial action maps, geometry-aware action representations, spatial affordance prediction, and structured spatial reasoning~\citep{zeng2021transporter,shridhar2022cliport,gervet2023act3d,ze20243ddiffusion,qu2025spatialvla,li2025bridgevla,huang2023voxposer,yuan2024robopoint}. These methods can improve the spatial expressiveness of the action interface itself. For continuous VLA action heads based on diffusion or flow matching~\citep{black2024pi0,bjorck2025gr00tn1,liu2024rdt}, an open challenge is how to dynamically select execution-relevant local geometry while maintaining a compact conditioning interface for action generation.

We offer a solution through \method, a State-Guided Spatial Alignment architecture that uses RGB-derived geometry and proprioceptive state to condition action generation. \method post-trains a Depth Anything V2 geometry branch with robot-domain RGB-D supervision, discards the depth head, and retains Geometry-Enhanced Post-Trained (GEP) features for policy conditioning. Proprioceptive state queries the GEP grid to produce compact geometry tokens for a flow-matching DiT action decoder~\citep{lipman2022flowmatching,peebles2023dit}; neither measured nor predicted depth maps are policy inputs. The complete recipe achieves \res{99.0}\%, \res{85.3}\%, and \res{78.8}\% average success on LIBERO, SimplerEnv-Fractal, and ALOHA. Its comparison with RGB-only includes additional RGB-D supervision; matched-supervision ablations support the combination of post-trained geometry and state-guided querying.

Our contributions are summarized below.
\begin{itemize}
    \item \textbf{State-Guided Spatial Alignment.} We introduce an architecture where a robot's proprioceptive state queries a geometry feature grid, extracting compact, state-conditioned geometry tokens to inform action generation.
    \item \textbf{Geometry-Enhanced Post-Trained feature.} We obtain RGB-derived GEP features by post-training a depth prediction model with robot-domain RGB-D supervision, then use encoder-side features for policy conditioning instead of raw depth predictions.
    \item \textbf{Simulation and real-world validation.} We evaluate \method on LIBERO, SimplerEnv-Fractal, and real-world ALOHA tasks using Isaac-GR00T N1.6-3B, with ablations examining geometry post-training and state-guided querying.
\end{itemize}

%===============================================================================
\section{Related Work}
\label{sec:related_work}

\paragraph{Vision--language--action policies.}
Recent VLA work adapts pretrained vision--language representations with robot demonstration data to build generalist policies for language-conditioned manipulation~\citep{brohan2023rt1,zitkovich2023rt2,oneill2023openx,kim2025openvla,black2024pi0,black2025pi05,bjorck2025gr00tn1}. Alongside scaling robot data, embodiments, and model capacity, this line has explored diverse action interfaces, including discrete action tokens, diffusion policies, and flow-matching decoders for continuous action chunks~\citep{chi2023diffusionpolicy,lipman2022flowmatching,liu2024rdt}. Recent predictive and generative world-model work further broadens the policy context through future-state prediction, goal-state generation, predictive rollouts, and imagination-based policy improvement~\citep{zhang2025dreamvla,cen2025worldvla,chen2025goalvla,yang2026rise,tian2026starry}. Together, these directions expand VLA policy learning across semantics, prediction, and action generation, while spatial guidance for fine-grained action decoding remains a complementary thread.

\paragraph{Spatial perception for VLAs.}
Recent work strengthens the VLM/perception side by giving visual-language representations stronger spatial content. SpatialVLM and RoboSpatial train VLMs for 2D/3D spatial understanding~\citep{chen2024spatialvlm,song2025robospatial}; 3D-CAVLA, DepthVLA, QDepth-VLA, and SG-VLA add depth-aware modules, 3D context, spatial grounding, or auxiliary depth supervision to VLA policies~\citep{bhat20253dcavla,yuan2025depthvla,li2025qdepthvla,tu2026sgvla}. These methods improve the information available before action decoding. \method instead uses RGB-D supervision to learn RGB-derived geometry features and does not use measured depth as a rollout input.

\paragraph{Spatial action generation.}
Complementary work makes the action interface itself spatial. Transporter and CLIPort predict spatial pick-place actions~\citep{zeng2021transporter,shridhar2022cliport}; PerAct, RVT, Act3D, and 3D Diffusion Policy use voxel, multi-view, point-cloud, or 3D feature representations for action prediction~\citep{shridhar2023peract,goyal2023rvt,gervet2023act3d,ze20243ddiffusion}. Recent VLA and affordance methods introduce adaptive action grids, 3D action experts, heatmap action outputs, value maps, spatial affordance prediction, or spatial action reasoning~\citep{qu2025spatialvla,sun2025geovla,li2025bridgevla,huang2023voxposer,yuan2024robopoint,lee2025molmoact}. \method shares the goal of spatially guided action generation, but uses proprioceptive-state queries to produce compact geometry tokens for a continuous DiT action head.

\paragraph{State-guided attention and spatial alignment.}
\method uses standard cross-attention and is related to latent-query architectures such as Perceiver, Q-Former, and Slot Attention~\citep{jaegle2021perceiver,li2023blip2,locatello2020slotattention}. Our contribution is not a new attention primitive. Instead, \method instantiates state-guided attention as a policy-conditioning mechanism, where robot state queries image-space geometry features to supply state-conditioned spatial cues to the action decoder.

%===============================================================================
\section{Method}
\label{sec:method}

We formulate \method in a language-conditioned VLA policy-learning setting. At each time step, the policy observes multi-view RGB images $I_t$, a language instruction $\ell$, and proprioceptive state $s_t$, and predicts an action chunk $A_t\in\R^{H\times d_a}$ with horizon $H$. The policy is trained from demonstrations $\Dpol$, while a separate robot-domain RGB-D dataset $\Ddep$ is used only for offline geometry post-training. During policy rollout, the policy conditions on RGB images, language, and robot state, not measured depth, point clouds, or depth maps. \method factorizes the decoder context into RGB-language semantic tokens and compact geometry tokens whose attention outputs depend on the current observation and robot state.

\begin{figure}[!t]
    \centering
    \includegraphics[width=\linewidth]{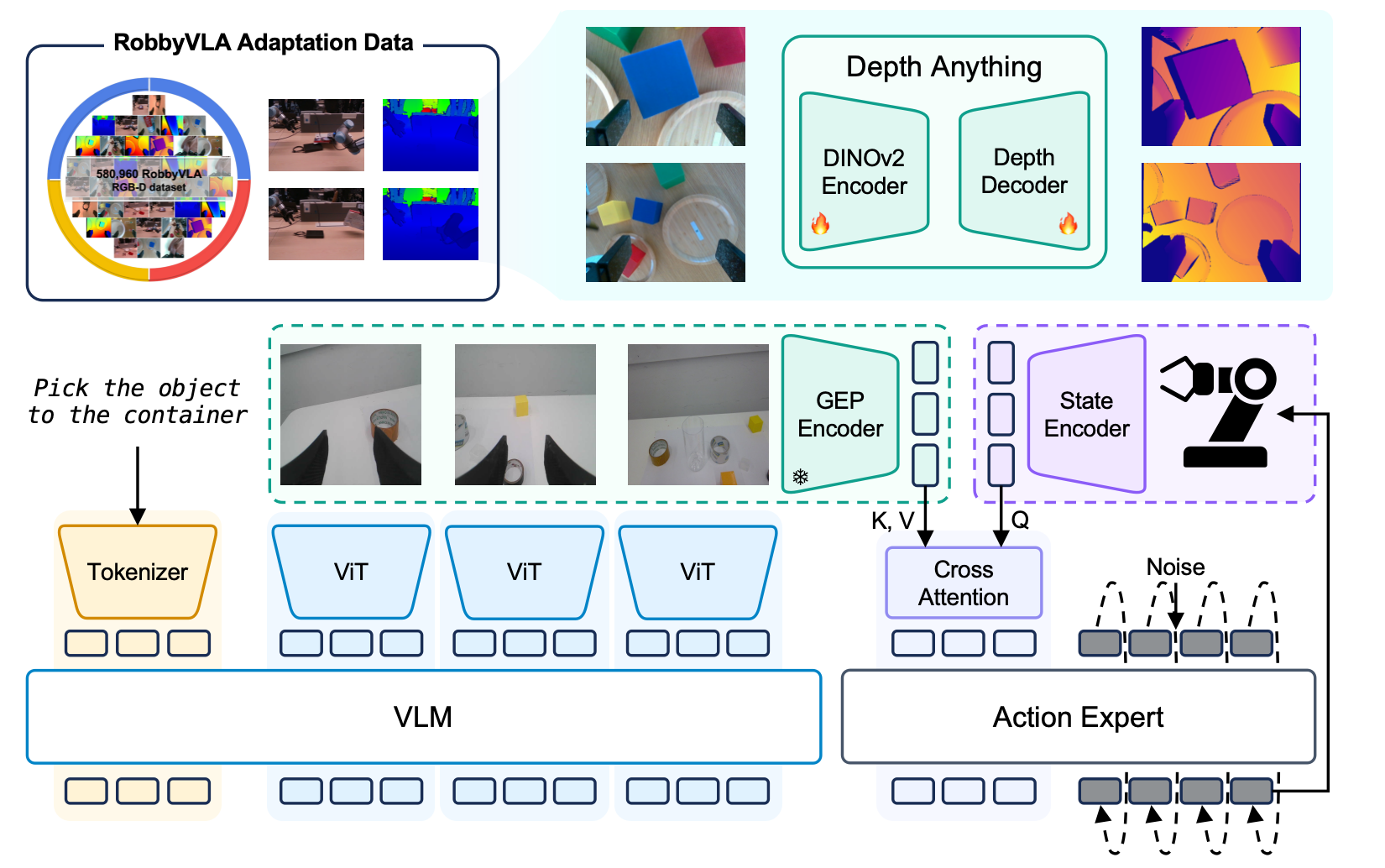}
    \caption{\textbf{GeoAlign overview.} Top: offline geometry post-training uses robot-domain RGB-D supervision, and the depth head is discarded afterward. Bottom: during policy training and rollout, a frozen RGB geometry branch produces image-space \gep features from RGB images, and proprioceptive-state queries extract compact geometry tokens that guide the Isaac-GR00T N1.6-3B DiT action head together with RGB-language tokens.}
    \label{fig:overview}
\end{figure}

\subsection{Overview}
\label{sec:method_overview}

Figure~\ref{fig:overview} illustrates the two-stage \method pipeline. Offline geometry post-training adapts a Depth Anything V2-Small model with robot-domain RGB-D supervision and then discards the depth head. During policy training and rollout, multi-view RGB and language produce semantic tokens $Z_t^{\mathrm{vlm}}$, while the frozen geometry branch maps the same RGB images to an image-space \gep feature grid. Proprioceptive-state query slots $Q_t$ cross-attend to this grid, producing compact geometry tokens concatenated with semantic tokens to guide the flow-matching DiT action decoder~\citep{peebles2023dit}.

\subsection{Geometry-Enhanced Post-Trained Feature Extraction}
\label{sec:gep_features}

We initialize the geometry branch from Depth Anything V2-Small~\citep{yang2024depthanythingv2} and post-train it on the RobbyVLA subset of LingBot-Depth~\citep{tan2026maskeddepth}: 580,960 paired RGB-D frames from Franka and UR7e manipulation. The encoder is post-trained once and then reused on LIBERO, Google Robot, and ALOHA. No LIBERO or SimplerEnv depth data or evaluated ALOHA demonstrations are used for this stage, and no new depth collection is required in these evaluated settings. Appendix~\ref{app:implementation} provides the full configuration.

During geometry post-training, given RGB image $I$ and metric depth $D$, the temporary depth head predicts $\hat{D}=E_{\mathrm{dep}}(I)$. We optimize valid pixels with the Scale-Invariant Logarithmic (SiLog) loss.

After post-training, we discard the depth prediction head and use the retained encoder-side descriptors as \gep features. Thus, depth supervision shapes the RGB-derived geometry representation, but the policy is conditioned on encoder-side \gep features rather than predicted depth maps. A linear projector maps reassembled features to $d_g=256$, yielding 5,476 spatial tokens per view, or $L_g=N\cdot 5476$ tokens across $N$ views. These tokens are augmented with 2D positional and view embeddings to form the geometry feature grid $\Phi_t^{\mathrm{geo}}\in\R^{B\times L_g\times d_g}$, preserving image-space spatial structure without explicit 3D reconstruction. Appendix~\ref{app:implementation} provides the full resolution and tokenization details.

\subsection{State-Guided Spatial Alignment}
\label{sec:routing}

The alignment module conditions geometry attention on the current robot state. We generate queries from proprioception, while RGB and language remain available through semantic tokens and RGB-derived geometry keys and values. This is a routing design choice rather than a demonstrated optimum. Proprioception also remains part of the inherited GR00T action-head input.

The state encoder maps $s_t$ to a single state embedding $h_t=\LN(E_s(s_t))\in\R^{B\times d_h}$. An MLP then generates $K=8$ query slots with a learned positional embedding $P^q$:
\begin{equation}
    Q_t=\reshape(W_q h_t, K, d_g)+P^q
    \in\R^{B\times K\times d_g}.
\end{equation}
These state-guided queries cross-attend to the geometry feature grid $\Phi_t^{\mathrm{geo}}$ with 8 attention heads, followed by a feed-forward block:
\begin{equation}
    \bar{G}_t = Q_t + \MHA(\LN(Q_t), \LN(\Phi_t^{\mathrm{geo}}), \LN(\Phi_t^{\mathrm{geo}})),
    \quad
    G_t = \bar{G}_t + \FFN(\LN(\bar{G}_t)).
\end{equation}
The attended geometry features are then projected to the VLA decoder dimension:
\begin{equation}
    Z_t^{\mathrm{geo}}=W_z\LN(G_t)\in\R^{B\times K\times d_v}.
\end{equation}
The output is 8 compact geometry tokens. A fixed state yields the same queries, but attention outputs remain scene-dependent because keys and values come from the current RGB-derived grid. We use $K=8$ throughout; precise-manipulation results establish its viability on these tasks, but we did not sweep $K$ or establish lossless compression. Appendix~\ref{app:diagnostics} provides qualitative attention diagnostics.

\subsection{Action Generation and Training}
\label{sec:decoder}

Following the GR00T action interface, the RGB-language backbone encodes RGB observations and language into semantic conditioning tokens $Z_t^{\mathrm{vlm}}=E_{\mathrm{vlm}}(I_t,\ell)\in\R^{B\times L_v\times d_v}$. We concatenate the compact geometry tokens to these VLM tokens:
\begin{equation}
    C_t=[Z_t^{\mathrm{vlm}};Z_t^{\mathrm{geo}}]\in\R^{B\times (L_v+8)\times d_v}.
\end{equation}
The decoder follows the GR00T flow-matching DiT action head~\citep{bjorck2025gr00tn1}. For a ground-truth action chunk $A_t$ and Gaussian noise $\epsilon\sim\mathcal{N}(0,I)$, we sample a flow time $\tau$ from the GR00T beta time sampler and construct $x_\tau=(1-\tau)\epsilon+\tau A_t$ with target velocity $v^\star=A_t-\epsilon$. The noised action chunk and discretized flow time are embedded as action tokens, concatenated with the encoded robot state and GR00T future tokens, and processed by the DiT while cross-attending to $C_t$. The decoder predicts the velocity on the action-token positions, denoted $\hat{v}_\theta$, and is trained with the same action-mask normalization used by the GR00T action head:
\begin{equation}
    \mathcal{L}_{\mathrm{policy}}=
    \E_{(I_t,\ell,s_t,A_t),\epsilon,\tau}
    \left[
    \frac{
    \sum_{i,j} m_{t,i,j}\left(\hat{v}_{\theta,i,j}-v^\star_{i,j}\right)^2
    }{
    \sum_{i,j} m_{t,i,j}
    }
    \right].
\end{equation}
Here $m_t$ is the GR00T action mask. The inherited 0.8 state-dropout probability applies to the encoded state entering the GR00T action head during policy training; it does not apply to the state used to generate geometry queries. At inference, we initialize from Gaussian noise and integrate the predicted velocity with $N_{\mathrm{ode}}$ Euler steps, executing the action before replanning.

\paragraph{Training and complexity.}
Training proceeds in two stages: post-train Depth Anything V2 on $\Ddep$ with $\mathcal{L}_{\mathrm{dep}}$, then discard the depth head and freeze the post-trained geometry encoder while training the projector, alignment module, and action decoder on $\Dpol$ with $\mathcal{L}_{\mathrm{policy}}$. The alignment module attends over the geometry grid but appends only 8 compact geometry tokens to the DiT conditioning context; Appendix~\ref{app:implementation} summarizes training and Appendix~\ref{app:diagnostics} reports runtime.

%===============================================================================
\section{Experiments}
\label{sec_experiments}

Our experiments evaluate the complete \method recipe on Isaac-GR00T N1.6-3B, examine its components through ablations, and test its utility for real-world geometry-critical manipulation.

\subsection{Experimental setup}
\label{sec_exp_setup}

\paragraph{Environments.}
We evaluate on LIBERO~\citep{liu2023libero} (four suites, 8,000 total rollouts), three shared SimplerEnv Google Robot Fractal task families~\citep{li2024simpler} (Pick Coke Can, Move Near, and Open/Close Drawer), and a real-world AgileX ALOHA platform~\citep{zhao2023aloha} (eight tabletop tasks). The real-world tasks cover geometry-critical settings, including transparent containers, annular tape rolls, occluded clear tape, and plate-relative placement. Appendix~\ref{app:detailed_results} provides full task descriptions and platform details.

\paragraph{Baselines.}
All controlled variants share the same Isaac-GR00T N1.6-3B backbone~\citep{bjorck2025gr00tn1} and evaluation protocol, differing only in the geometry branch and state-guided alignment module. \textbf{RGB-only backbone} is the base Isaac-GR00T N1.6-3B backbone. \textbf{w/o post-training} uses unadapted DA-V2 encoder features with state-guided alignment. \textbf{w/o spatial querying} keeps the post-trained \gep grid but replaces spatial cross-attention with global average pooling into static geometry tokens. \textbf{w/o state queries} keeps spatial cross-attention but uses learned query slots instead of proprioceptive-state-generated queries. \textbf{w/ unfrozen encoder} updates the geometry encoder during policy training. \textbf{\method} is the full model. We additionally compare against $\pi_{0.5}$~\citep{black2025pi05} on real-world tasks. Appendix~\ref{app:detailed_results} provides the full baseline specifications.

\paragraph{Metrics.}
We report success rates. Fractal uses the unweighted average over three task families, and ALOHA averages eight tasks with 20 trials each. Aggregate LIBERO and ALOHA results include descriptive rollout-level Wilson 95\% confidence intervals; these do not measure training-seed variability or uncertainty in paired differences.

\paragraph{Auxiliary-data control.}
\method uses external RobbyVLA RGB-D supervision that RGB-only and $\pi_{0.5}$ do not receive. Their comparisons evaluate the complete recipe, not an architecture-only effect. The w/o spatial querying and w/o state queries variants use the same post-trained GEP encoder and auxiliary data as \method, enabling comparisons of geometry conditioning under matched supervision. We did not train RGB-only or $\pi_{0.5}$ with an alternative RobbyVLA objective.

\paragraph{Implementation.}
All geometry-feature variants use RGB observations as visual input during evaluation and do not use measured depth inputs. The model uses an Isaac-GR00T N1.6-3B backbone~\citep{bjorck2025gr00tn1}\footnote{Backbone checkpoint: \url{https://huggingface.co/nvidia/GR00T-N1.6-3B}.} with Eagle-Block2A-2B VLM features, \gep features, and a 32-layer DiT action head~\citep{peebles2023dit}. All variants share the policy-training protocol; Appendix~\ref{app:implementation} reports the full configuration.

\begin{figure}[t]
    \centering
    \includegraphics[width=\linewidth]{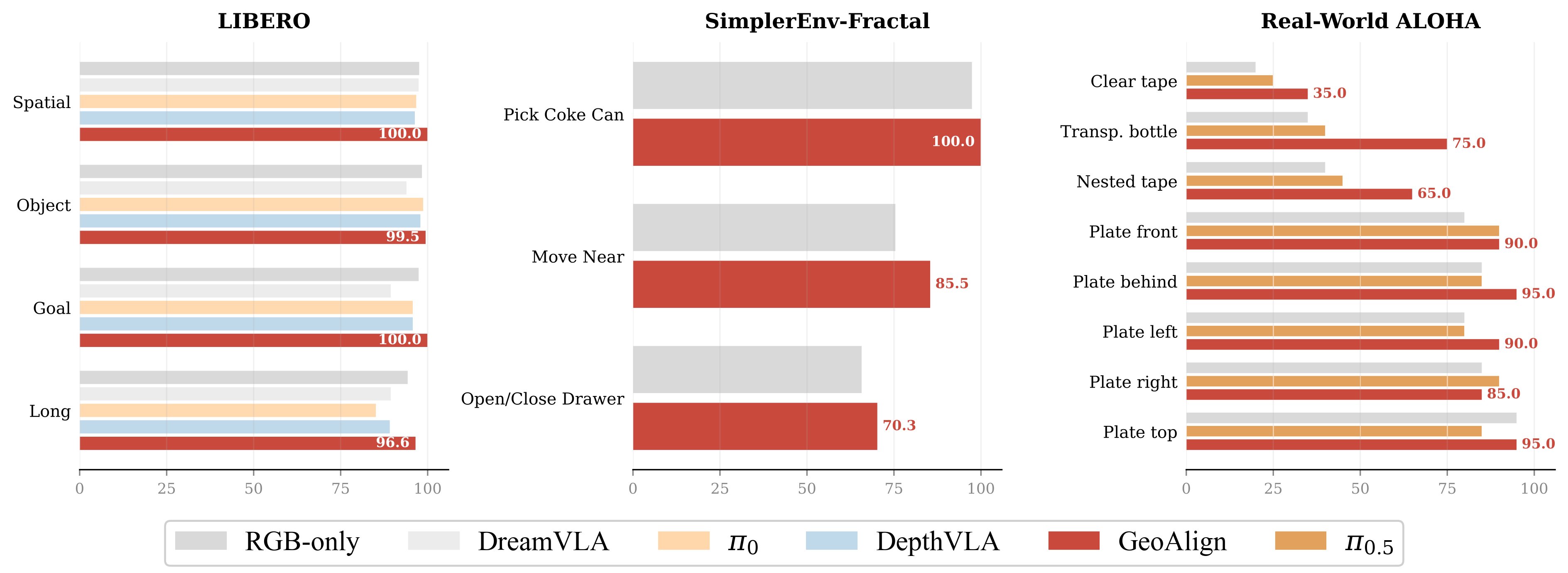}
    \caption{\textbf{Evaluation overview.} Success rates on LIBERO, SimplerEnv-Fractal, and real-world ALOHA; \method is shown in red.}
    \label{fig:overview_results}
\end{figure}

\subsection{Simulation Results on LIBERO and SimplerEnv-Fractal}
\label{sec_simulation_results}

Figure~\ref{fig:overview_results} summarizes the three evaluation settings. \method achieves \res{99.0}\% average success across the four LIBERO suites (7,922/8,000; 95\% CI [98.8, 99.2]), compared with \res{97.0}\% for RGB-only (7,759/8,000; [96.6, 97.3]). Spatial improves from \res{97.65}\% to \res{100.0}\%, and Long from \res{94.35}\% to \res{96.6}\%. These gains include auxiliary RGB-D post-training. Appendix Table~\ref{tab:libero_public} provides published-result context; our within-backbone comparisons share policy demonstrations, policy-training settings, and evaluation protocol.

Appendix Table~\ref{tab:simplerenv_fractal} reports results on the three shared SimplerEnv-Fractal Visual Matching task families, Pick Coke Can, Move Near, and Open/Close Drawer. \method achieves \res{100.0}\%, \res{85.5}\%, and \res{70.3}\% success, corresponding to a \res{85.3}\% unweighted average (\res{+5.7} percentage points over RGB-only). Public-context rows are included only to situate the result; the controlled Fractal comparison is the gain over RGB-only under the same policy setup.

\subsection{Ablation Analysis}
\label{sec_ablation}

Appendix Table~\ref{tab:ablation_results} examines the geometry encoder and query source under the same backbone, policy demonstrations, policy-training settings, evaluation protocol, and checkpoint selection. Auxiliary RGB-D supervision varies only with the post-training condition.

\paragraph{Robot-domain geometry post-training.} The w/o post-training variant replaces post-trained \gep features with unadapted DA-V2 encoder features and drops average LIBERO success from \res{99.0}\% to \res{95.9}\%. The w/ unfrozen encoder variant also underperforms the frozen \method variant (\res{95.93}\%), suggesting that preserving the post-trained representation is preferable.

\paragraph{Proprioceptive state querying.} With the same post-trained encoder and auxiliary supervision, global pooling (w/o spatial querying, \res{91.6}\%) and learned queries (w/o state queries, \res{96.2}\%) both underperform \method (\res{99.0}\%). Learned versus state-generated queries differ only in query generation; global pooling additionally removes spatial cross-attention. No partial variant exceeds RGB-only (\res{97.0}\%) in this run. The evidence therefore supports combining robot-domain post-training, preservation of the spatial grid, and state-guided querying; it does not show that each component independently improves RGB-only.

\subsection{Real-World ALOHA Evaluation and Attention Behavior}
\label{sec_real_world_results}

We evaluate eight tasks on AgileX ALOHA, covering transparent containers, annular tape-roll alignment, occluded clear tape, and plate-relative placement. Table~\ref{tab:aloha_results} reports per-task success and aggregate uncertainty. \method achieves \res{78.8}\% average success, an observed 13.8-percentage-point gain over RGB-only (\res{65.0}\%), and is numerically above $\pi_{0.5}$ (\res{67.5}\%). The aggregate intervals for \method and $\pi_{0.5}$ overlap. We make no significance claim for their difference or for individual 20-trial tasks; the RGB-only comparison evaluates the complete recipe including auxiliary RGB-D supervision.

\begin{table}[H]
\centering
\caption{\textbf{Real-world ALOHA deployment.} Success rates (\%) over 20 trials per task. Aggregate rows pool 160 trials; intervals are Wilson 95\% CIs. Per-task differences are descriptive.}
\label{tab:aloha_results}
\begingroup
\small
\setlength{\tabcolsep}{4pt}
\begin{tabular}{@{}lccc@{}}
\toprule
Task & RGB-only & $\pi_{0.5}$ & GeoAlign \\
\midrule
Clear tape & \res{20.0} & \res{25.0} & \res{35.0} \\
Transparent bottle & \res{35.0} & \res{40.0} & \res{75.0} \\
Tape-roll insertion & \res{40.0} & \res{45.0} & \res{65.0} \\
Plate front & \res{80.0} & \res{90.0} & \res{90.0} \\
Plate behind & \res{85.0} & \res{85.0} & \res{95.0} \\
Plate left & \res{80.0} & \res{80.0} & \res{90.0} \\
Plate right & \res{85.0} & \res{90.0} & \res{85.0} \\
Plate top & \res{95.0} & \res{85.0} & \res{95.0} \\
\midrule
Average & \res{65.0} & \res{67.5} & \res{78.8} \\
Successes / trials & 104/160 & 108/160 & 126/160 \\
95\% CI & [57.3, 72.0] & [59.9, 74.3] & [71.8, 84.4] \\
\bottomrule
\end{tabular}
\par\smallskip
\footnotesize
\emph{Notes.} Tape-roll insertion denotes placing the small tape roll inside the large tape roll.
\endgroup
\end{table}

\begin{figure}[t]
    \centering
    \includegraphics[width=\linewidth]{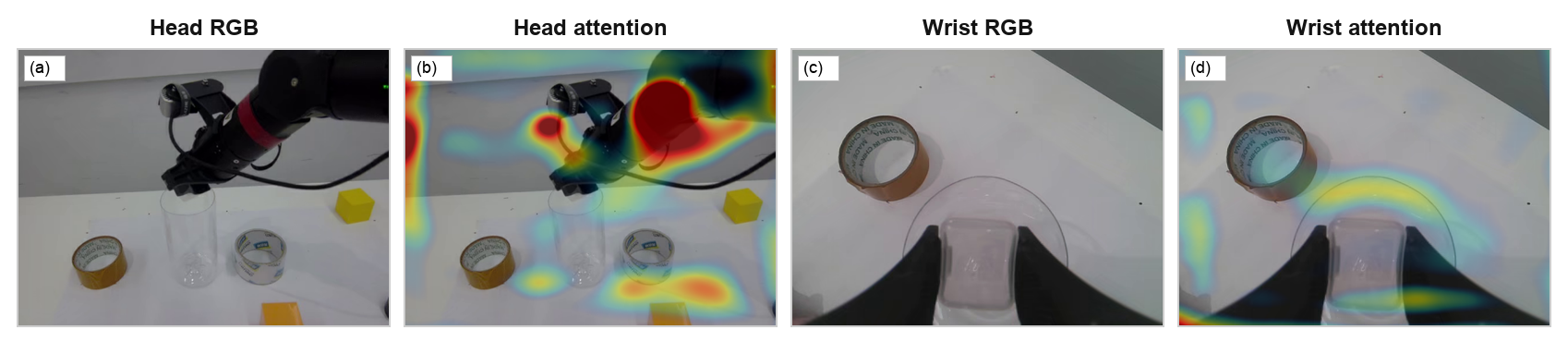}
    \caption{\textbf{State-guided geometry attention during real-world rollout.} RGB observations and geometry-query attention maps for a transparent-container task. The model is not trained with attention annotations; these maps are shown as qualitative diagnostics.}
    \label{fig:attention_main}
\end{figure}

Figure~\ref{fig:attention_main} visualizes geometry-query attention during a transparent-container rollout. Geometry queries are optimized for action prediction rather than segmentation, so attention need not peak on the named object. Spatial cues may also involve the end effector, container rim, object contours, occlusion boundaries, or clearance. These possibilities motivate a qualitative reading of the maps; they do not identify the cause of a particular activation. Attention remains diffuse in places and can activate image boundaries. Without region or phase annotations, the maps do not establish object localization or manipulation-phase switching. Appendix~\ref{app:diagnostics} provides the full multi-view visualization.

\paragraph{State-only routing and scene dependence.}
Proprioceptive state represents the arm and gripper configuration at the current control step. The task instruction usually stays fixed during a rollout, while that configuration changes as the robot acts; we therefore use state as the geometry-routing signal. Language and RGB semantics still condition the action decoder through the semantic tokens. A fixed state produces the same queries, but their attention outputs remain scene-dependent because keys and values come from the current RGB-derived \gep grid. Under matched RobbyVLA supervision and the same post-trained encoder, state-generated queries achieve \res{99.0}\% on LIBERO versus \res{96.2}\% for learned slots. This supports state routing within the tested recipe; we have not evaluated joint state-language or state-visual query generation.

\paragraph{Token budget and depth requirements.}
Each of the eight queries attends to the full $N\times5{,}476$ spatial-token grid, rather than eight sampled positions. The budget limits the number of geometry tokens appended to the decoder. The precise-manipulation results support this setting on the evaluated tasks; we did not sweep the budget or establish lossless compression. The encoder is post-trained once on Franka/UR7e RobbyVLA data and reused on LIBERO, Google Robot, and ALOHA without collecting target-platform depth in these evaluations.

\begin{figure}[t]
    \centering
    \includegraphics[width=\linewidth]{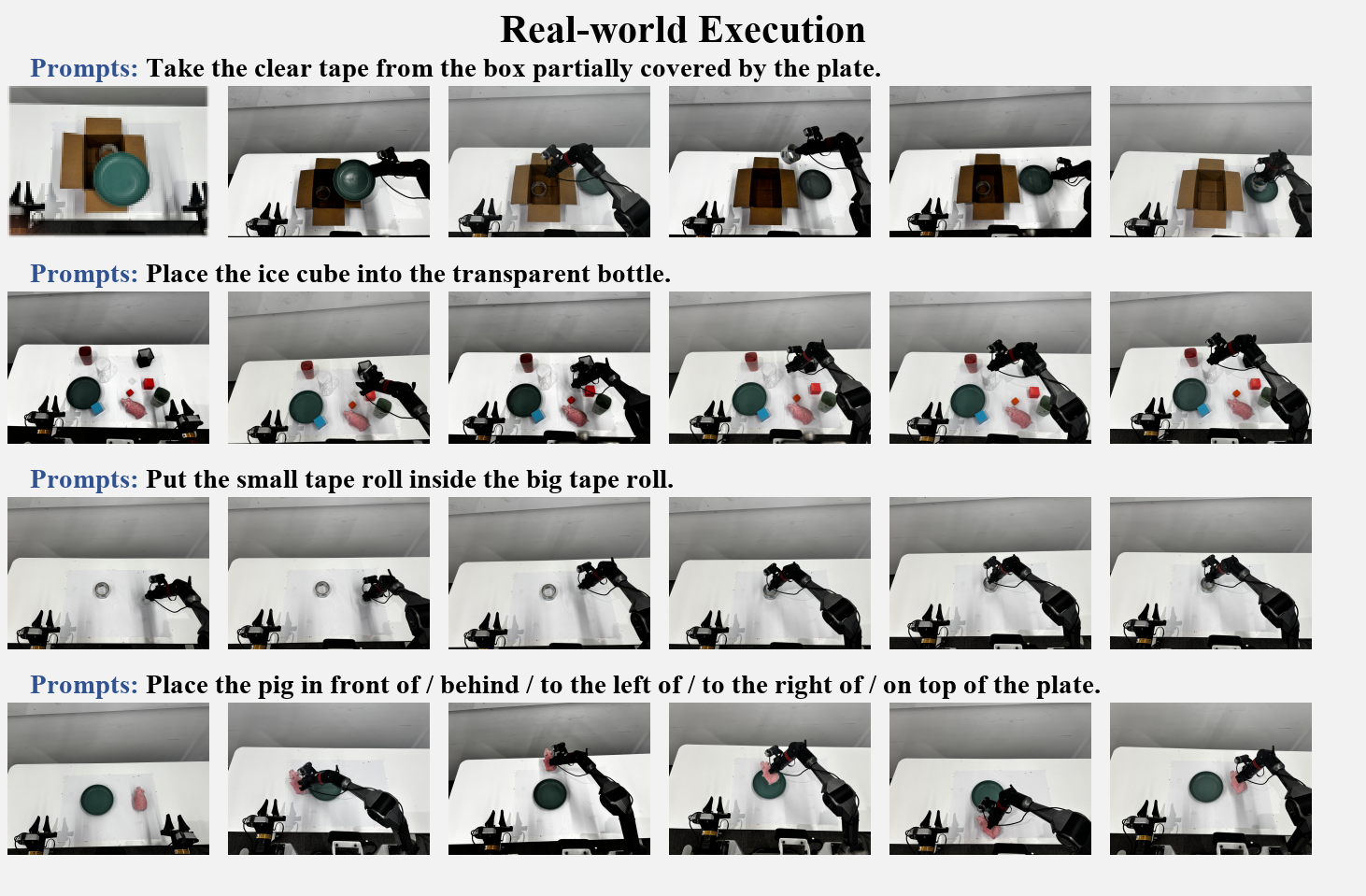}
    \caption{\textbf{Real-world execution examples.} Representative ALOHA rollouts for clear-tape retrieval, transparent-bottle insertion, ring-in-ring placement, and plate-relative placement.}
    \label{fig:real_world_execution}
\end{figure}

\FloatBarrier
%===============================================================================
\section{Limitations}
\label{sec:limitations}

\method requires external RGB-D post-training data that RGB-only and $\pi_{0.5}$ do not use; its gains over those baselines cannot be attributed to architecture alone. We did not evaluate alternative uses of the same auxiliary data or another VLA backbone. Transfer of the frozen encoder across the evaluated platforms does not guarantee transfer to arbitrary embodiments, cameras, or workspaces. We did not sweep the eight-token budget or compare state-only queries with state-language or state-visual queries. Attention maps provide qualitative diagnostics without annotated localization or phase evidence. The method also lacks explicit collision, reachability, and contact constraints and persistent scene memory. Broader deployment may require better data coverage, calibrated multi-view geometry, spatial memory, and contact or force feedback.

%===============================================================================
\section{Conclusion}
\label{sec:conclusion}

We presented \method, which combines offline robot-domain geometry post-training with state-generated queries over an RGB-derived feature grid. Eight geometry tokens condition a flow-matching action decoder, and measured or predicted depth maps are never policy inputs. On Isaac-GR00T N1.6-3B, the complete recipe improves observed success on LIBERO, three SimplerEnv-Fractal task families, and eight ALOHA tasks. Matched-supervision ablations support combining the post-trained encoder, spatial grid, and state queries. The real-world result is an observed aggregate gain over RGB-only and a numerically higher average than $\pi_{0.5}$; uncertainty and auxiliary-data differences limit stronger comparisons.

%===============================================================================
% The acknowledgments are automatically included only in the final and preprint versions of the paper.
\acknowledgments{This work was supported by the National Natural Science Foundation of China (Grant No. 62373242). We thank the anonymous reviewers and the area chair for their constructive feedback.}

%===============================================================================
\clearpage
\FloatBarrier
% no \bibliographystyle is required, since corl_2026 automatically uses corlabbrvnat.bst.
\bibliography{references}  % .bib

%===============================================================================
\clearpage
\appendix
\section*{Appendix}
\addcontentsline{toc}{section}{Appendix}

\section{Detailed Results}
\label{app:detailed_results}

\subsection{Experimental Setup}

\paragraph{Environments.}
In simulation, we use LIBERO~\citep{liu2023libero} (four suites: Spatial, Object, Goal, and Long; 10 tasks per suite; 200 rollouts per task, totalling 8,000 rollouts) and three shared SimplerEnv Google Robot / Fractal task families~\citep{li2024simpler}: Pick Coke Can, Move Near, and Open/Close Drawer. In the real world, we evaluate on a stationary bimanual AgileX ALOHA platform~\citep{zhao2023aloha} with eight tabletop tasks: taking clear tape from a box partially covered by a plate, placing an ice cube into a transparent bottle, placing a small tape roll inside a big tape roll, and five plate-relative placement tasks (front, behind, left, right, and on top). The real-world props are deliberately geometry-critical: transparent objects, clear tape, annular tape rolls, and plate-relative targets are easy to name semantically but require local geometry, clearance, and relative-position reasoning for successful execution. These tasks isolate local geometric execution rather than mobile navigation.

\paragraph{Baselines.}
All controlled variants share the same Isaac-GR00T N1.6-3B backbone~\citep{bjorck2025gr00tn1}, backbone initialization, demonstration data, task splits, evaluation seeds, rollout protocol, RGB observations, language instructions, proprioceptive state, action representation, decoder, training trajectories, training steps, batch size, optimizer settings, and checkpoint selection protocol. \textbf{RGB-only backbone} is the base Isaac-GR00T N1.6-3B backbone and serves as the controlled RGB-only baseline. \textbf{w/o post-training} uses unadapted DA-V2 encoder features with state-guided alignment, testing whether generic monocular depth priors suffice without robot-domain geometry post-training. \textbf{w/o spatial querying} keeps the post-trained \gep grid but replaces spatial cross-attention with global average pooling into static geometry tokens, testing whether simply adding post-trained geometry features is sufficient without spatial querying. \textbf{w/o state queries} keeps spatial cross-attention but uses learned query slots instead of proprioceptive-state-generated queries, isolating the value of proprioceptive querying. \textbf{w/ unfrozen encoder} updates the geometry encoder during policy training, testing whether the post-trained geometry representation should remain frozen. \textbf{\method} is the full model. For real-world deployment, we additionally compare against $\pi_{0.5}$~\citep{black2025pi05}, using the same collected training trajectories, camera streams, action space, and proprioceptive inputs.

\subsection{LIBERO Results}

\begin{table}[htbp]
\centering
\caption{LIBERO success rate (\%). Public baseline rows are copied from the DepthVLA LIBERO comparison table~\citep{yuan2025depthvla}; per-method citations indicate method provenance, not the source of the transcribed numbers.}
\label{tab:libero_public}
\begingroup
\small
\setlength{\tabcolsep}{4pt}
\begin{tabular}{@{}lccccc@{}}
\toprule
Method & Spatial & Object & Goal & Long & Avg. \\
\midrule
Octo-Base~\citep{ghosh2024octo} & 78.9 & 85.7 & 84.6 & 51.1 & 75.1 \\
OpenVLA~\citep{kim2025openvla} & 84.7 & 88.4 & 79.2 & 53.7 & 76.5 \\
SpatialVLA~\citep{qu2025spatialvla} & 88.2 & 89.9 & 78.6 & 55.5 & 78.1 \\
CoT-VLA~\citep{zhao2025cotvla} & 81.5 & 91.6 & 87.6 & 69.0 & 83.9 \\
MolmoAct~\citep{lee2025molmoact} & 87.0 & 95.4 & 87.6 & 77.2 & 86.6 \\
DreamVLA~\citep{zhang2025dreamvla} & 97.5 & 94.0 & 89.5 & 89.5 & 92.6 \\
$\pi_0$~\citep{black2024pi0} & 96.8 & 98.8 & 95.8 & 85.2 & 94.2 \\
DepthVLA~\citep{yuan2025depthvla} & 96.4 & 98.0 & 95.8 & 89.2 & 94.9 \\
\textbf{GeoAlign} & \res{100.0} & \res{99.5} & \res{100.0} & \res{96.6} & \res{99.0} \\
\bottomrule
\end{tabular}
\endgroup
\end{table}

\method achieves \res{99.0}\% average success across the four LIBERO suites. With 2,000 evaluation rollouts per suite, this corresponds to 7,922 successes out of 8,000 rollouts, compared with 7,759/8,000 for RGB-only. Rollout-level Wilson 95\% confidence intervals are \res{99.0}\% [\res{98.8}, \res{99.2}] for \method and \res{97.0}\% [\res{96.6}, \res{97.3}] for RGB-only. The largest observed gains are on Spatial (\res{100.0}\% vs. \res{97.65}\%) and Long (\res{96.6}\% vs. \res{94.35}\%). These results evaluate the complete recipe, including auxiliary geometry post-training.

Published LIBERO numbers in Table~\ref{tab:libero_public} provide context rather than fully controlled comparisons. Our within-backbone comparisons share policy demonstrations, policy-training settings, and evaluation protocol, while auxiliary RGB-D supervision differs for RGB-only and w/o post-training.

\subsection{SimplerEnv-Fractal Results}

\begin{table}[htbp]
\centering
\caption{SimplerEnv-Fractal Google Robot results on shared Visual Matching task families. Controlled rows are evaluated by us under the same policy setup; public-context rows are transcribed from the cited source tables.}
\label{tab:simplerenv_fractal}
\begingroup
\scriptsize
\setlength{\tabcolsep}{2pt}
\begin{tabular}{@{}llccc@{}}
\toprule
Method / setting & Source & \makecell{Pick Coke\\Can} & Move Near & \makecell{Open/Close\\Drawer} \\
\midrule
RGB-only baseline (ours) & ours & \res{97.5} & \res{75.5} & \res{65.8} \\
\textbf{GeoAlign (ours)} & ours & \res{100.0} & \res{85.5} & \res{70.3} \\
RT-1-X~\citep{oneill2023openx} & MolmoAct Tbl. 1~\citep{lee2025molmoact} & 56.7 & 31.7 & 59.7 \\
RT-2-X~\citep{zitkovich2023rt2} & MolmoAct Tbl. 1~\citep{lee2025molmoact} & 78.7 & 77.9 & 25.0 \\
OpenVLA~\citep{kim2025openvla} & MolmoAct Tbl. 1~\citep{lee2025molmoact} & 16.3 & 46.2 & 35.6 \\
Magma~\citep{yang2025magma} & MolmoAct Tbl. 1~\citep{lee2025molmoact} & 56.0 & 65.4 & 83.7 \\
GR00T-N1.5~\citep{bjorck2025gr00tn1} & MolmoAct Tbl. 1~\citep{lee2025molmoact} & 69.3 & 68.7 & 35.8 \\
SpatialVLA~\citep{qu2025spatialvla} & MolmoAct Tbl. 1~\citep{lee2025molmoact} & 81.0 & 69.6 & 59.3 \\
MolmoAct, zero-shot~\citep{lee2025molmoact} & MolmoAct Tbl. 1~\citep{lee2025molmoact} & 71.3 & 73.8 & 66.5 \\
MolmoAct, fine-tuned~\citep{lee2025molmoact} & MolmoAct Tbl. 1~\citep{lee2025molmoact} & 77.7 & 77.1 & 60.0 \\
CogACT~\citep{li2025cogact} & CogACT Table 1~\citep{li2025cogact} & 91.3 & 85.0 & 71.8 \\
\bottomrule
\end{tabular}
\endgroup
\end{table}

For the controlled rows, Open/Close Drawer averages the Open drawer and Close drawer tasks, following the Open/Close Drawer convention used in published Fractal reporting. Public-context rows are transcribed from the cited source tables; per-method citations indicate method provenance when available, while the Source column identifies where the numbers were copied from. These rows are included as context rather than strict head-to-head comparisons because training data, checkpoints, and evaluation infrastructure differ across publications. In this public-context comparison, \method is numerically above the SpatialVLA and MolmoAct rows on all three task families, strongest among the listed rows on Pick Coke Can and Move Near, close to CogACT on Open/Close Drawer, and below the Magma row copied from MolmoAct on Open/Close Drawer.

\subsection{Ablation Analysis}

\begin{table}[htbp]
\centering
\caption{Spatially enhanced VLA mechanism comparison. Rows summarize SpatialVLA~\citep{qu2025spatialvla}, GeoVLA~\citep{sun2025geovla}, 3D-CAVLA~\citep{bhat20253dcavla}, DepthVLA~\citep{yuan2025depthvla}, and QDepth-VLA~\citep{li2025qdepthvla}. This table is qualitative and intentionally avoids unverified numeric results for methods whose LIBERO splits were not audited here.}
\label{tab:depth_vla_comparison}
\begingroup
\scriptsize
\setlength{\tabcolsep}{3pt}
\resizebox{\linewidth}{!}{%
\begin{tabular}{@{}lcccccl@{}}
\toprule
\multirow{2}{*}{Method} & \multicolumn{4}{c}{Spatial source} & \multirow{2}{*}{\makecell{Grid\\kept}} & \multirow{2}{*}{Guidance} \\
 & \makecell{3D coords.\\or point cloud} & \makecell{Depth map\\or 3D ROI} & \makecell{Geometry\\branch} & \makecell{Quantized\\depth tokens} & & \\
\midrule
SpatialVLA & \cmark & \xmark & \xmark & \xmark & \xmark & Ego3D PE \\
GeoVLA & \cmark & \xmark & \xmark & \xmark & \xmark & 3D action expert \\
3D-CAVLA & \cmark & \cmark & \xmark & \xmark & \pmark & ROI pooling \\
DepthVLA & \xmark & \xmark & \cmark & \xmark & \xmark & shared attention \\
QDepth-VLA & \xmark & \xmark & \xmark & \cmark & \xmark & aux. depth pred. \\
\textbf{GeoAlign} & \xmark & \xmark & \textbf{\cmark} & \xmark & \textbf{\cmark} & \textbf{state query} \\
\bottomrule
\end{tabular}%
}
\par\smallskip
\begin{minipage}{0.92\linewidth}
\footnotesize
\emph{Notes.} \cmark: used/retained; \xmark: not used/retained; \pmark: partially retained region-level spatial context. Grid kept means an image-space geometry feature grid is retained for policy conditioning, not an action discretization grid. Geometry branch covers DepthVLA's depth transformer and GeoAlign's RGB-derived \gep feature branch; GeoAlign does not use measured depth during policy rollout.
\end{minipage}
\endgroup
\end{table}

\begin{table}[htbp]
\centering
\caption{LIBERO ablations of RGB-derived geometry conditioning. All variants use the same Isaac-GR00T N1.6-3B VLA backbone~\citep{bjorck2025gr00tn1} and evaluation protocol.}
\label{tab:ablation_results}
\begingroup
\small
\setlength{\tabcolsep}{4pt}
\begin{tabular}{@{}lccccc@{}}
\toprule
Variant & Spatial & Object & Goal & Long & Avg. \\
\midrule
RGB-only backbone & \res{97.65} & \res{98.45} & \res{97.5} & \res{94.35} & \res{97.0} \\
w/o post-training & \res{91.35} & \res{99.4} & \res{98.0} & \res{95.0} & \res{95.9} \\
w/o spatial querying & \res{90.05} & \res{96.5} & \res{92.5} & \res{87.5} & \res{91.6} \\
w/o state queries & \res{95.8} & \res{99.3} & \res{97.9} & \res{91.7} & \res{96.2} \\
w/ unfrozen encoder & \res{97.10} & \res{99.60} & \res{97.42} & \res{89.60} & \res{95.93} \\
\textbf{GeoAlign} & \res{100.0} & \res{99.5} & \res{100.0} & \res{96.6} & \res{99.0} \\
\bottomrule
\end{tabular}
\par\smallskip
\footnotesize
\emph{Notes.} ``w/o post-training'' uses unadapted DA-V2 features. ``w/o spatial querying'' replaces \gep cross-attention with global average pooling. ``w/o state queries'' uses learned query slots rather than proprioceptive-state-generated queries. ``w/ unfrozen encoder'' updates the geometry encoder during policy training.
\endgroup
\end{table}

All variants share the same VLA backbone and training protocol. The w/o state queries variant and \method are identical in the post-trained geometry encoder, projector, cross-attention block, number of query tokens, normalization layers, decoder, optimizer, training steps, and checkpoint selection; the only difference is whether the query slots are learned parameters or generated from proprioceptive state.

\paragraph{Geometry post-training matters.}
Comparing w/o post-training against \method---both use state-guided queries, but differ in whether the geometry features come from unadapted DA-V2 encoder features or robot-domain post-trained \gep features---\method improves the average from \res{95.9}\% to \res{99.0}\%, with the largest gain on Spatial (\res{91.35}\% vs.\ \res{100.0}\%). This controlled ablation supports the value of robot-domain geometry post-training when local geometry must be selected from cluttered manipulation scenes. The w/ unfrozen encoder variant reaches \res{95.93}\% average, below the frozen \method variant, suggesting that preserving the post-trained representation is preferable.

\paragraph{State-guided querying matters.}
With matched post-trained geometry features and auxiliary supervision, w/o spatial querying reaches \res{91.6}\% average success and w/o state queries reaches \res{96.2}\%, compared with \res{99.0}\% for \method. The latter comparison changes only query generation, with observed gains on Spatial (\res{95.8}\% to \res{100.0}\%) and Long (\res{91.7}\% to \res{96.6}\%). Global pooling additionally removes the spatial attention block. No partial variant exceeds RGB-only (\res{97.0}\%) in this run, so these ablations support the combined design rather than independent improvements from each component.

\subsection{Real-World ALOHA Results}

Each task in Table~\ref{tab:aloha_results} uses 20 evaluation trials. The average row aggregates 160 trials; Wilson 95\% confidence intervals are [57.3, 72.0] for RGB-only, [59.9, 74.3] for $\pi_{0.5}$, and [71.8, 84.4] for \method.

We evaluate on a stationary bimanual AgileX ALOHA platform with one high camera and two wrist cameras. The task set is designed so that object identity alone is insufficient: transparent containers require estimating the bottle mouth and interior, tape rolls require annular alignment, clear tape under partial occlusion requires local graspable-geometry recovery, and plate-relative placement requires releasing the object in the instructed spatial region. The first three manipulation tasks use 40 collected training trajectories each, while the five plate-relative placement tasks use only 10 collected training trajectories each, probing whether the policy can learn executable spatial relations from limited real demonstrations. Each policy is evaluated for 20 real-world trials per task. Object and target poses are randomized within an approximately $20\text{ cm}\times20\text{ cm}$ reachable tabletop region around the nominal task pose, with yaw randomized when it does not invalidate the task; cameras, robot base, and lighting are fixed. Because each real-world task uses 20 trials, the comparison to $\pi_{0.5}$ should be interpreted with the corresponding binomial uncertainty in mind; the main real-world conclusion is that \method improves on average over the controlled RGB-only baseline and is numerically above $\pi_{0.5}$ on average in our setup.

\begin{figure}[h]
    \centering
    \includegraphics[width=0.92\linewidth]{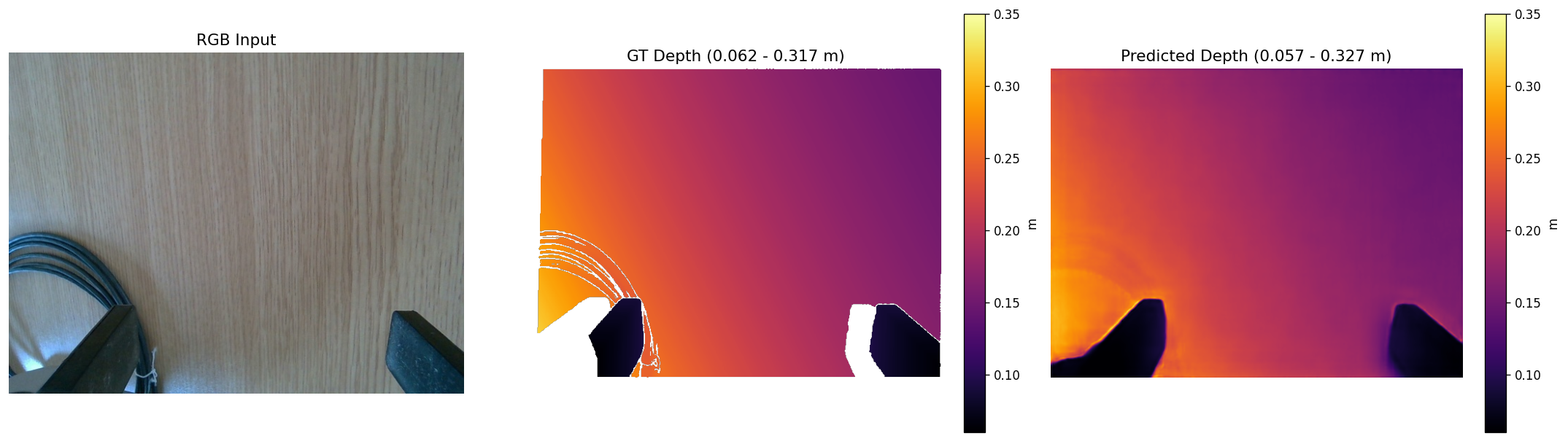}
    \caption{\textbf{GEP feature diagnostic.} RGB input, measured depth, and predicted depth are shown with the same crop and scale range used for the RobbyVLA geometry-feature validation audit.}
    \label{fig:depth_rgb_gt_pred}
\end{figure}

\begin{figure}[h]
    \centering
    \begin{minipage}[t]{0.32\linewidth}
        \centering
        \includegraphics[width=\linewidth]{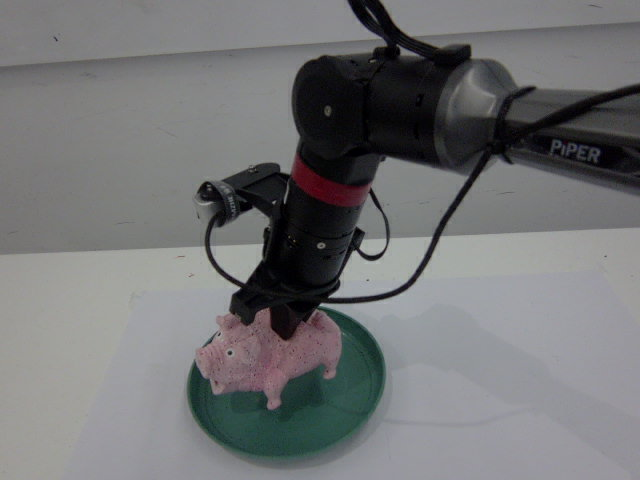}
        \par\smallskip
        \small (a) High view
    \end{minipage}
    \hfill
    \begin{minipage}[t]{0.32\linewidth}
        \centering
        \includegraphics[width=\linewidth]{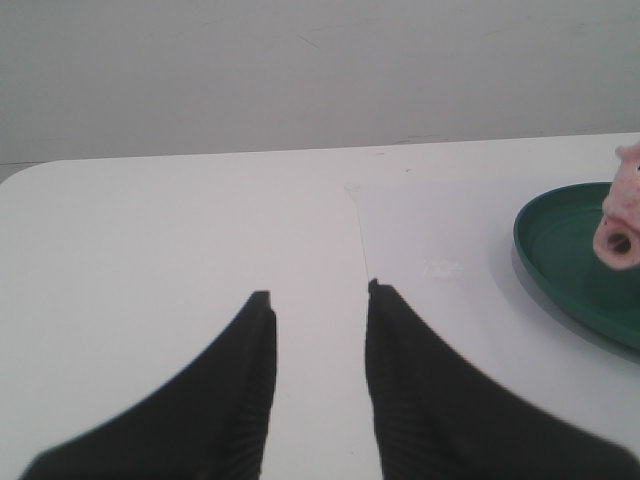}
        \par\smallskip
        \small (b) Left wrist
    \end{minipage}
    \hfill
    \begin{minipage}[t]{0.32\linewidth}
        \centering
        \includegraphics[width=\linewidth]{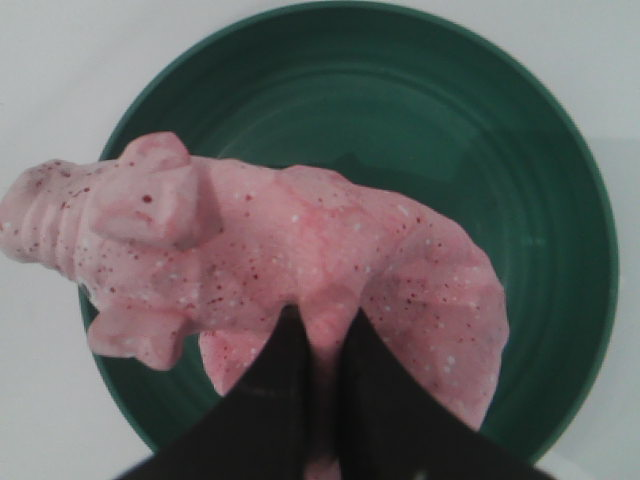}
        \par\smallskip
        \small (c) Right wrist
    \end{minipage}
    \caption{\textbf{Real-world AgileX ALOHA setup.} Representative camera views for tasks involving container placement, object transfer, and plate-relative spatial relations.}
    \label{fig:aloha_qualitative}
\end{figure}

\FloatBarrier
\section{Diagnostics}
\label{app:diagnostics}

\begin{figure}[h]
    \centering
    \includegraphics[width=\linewidth]{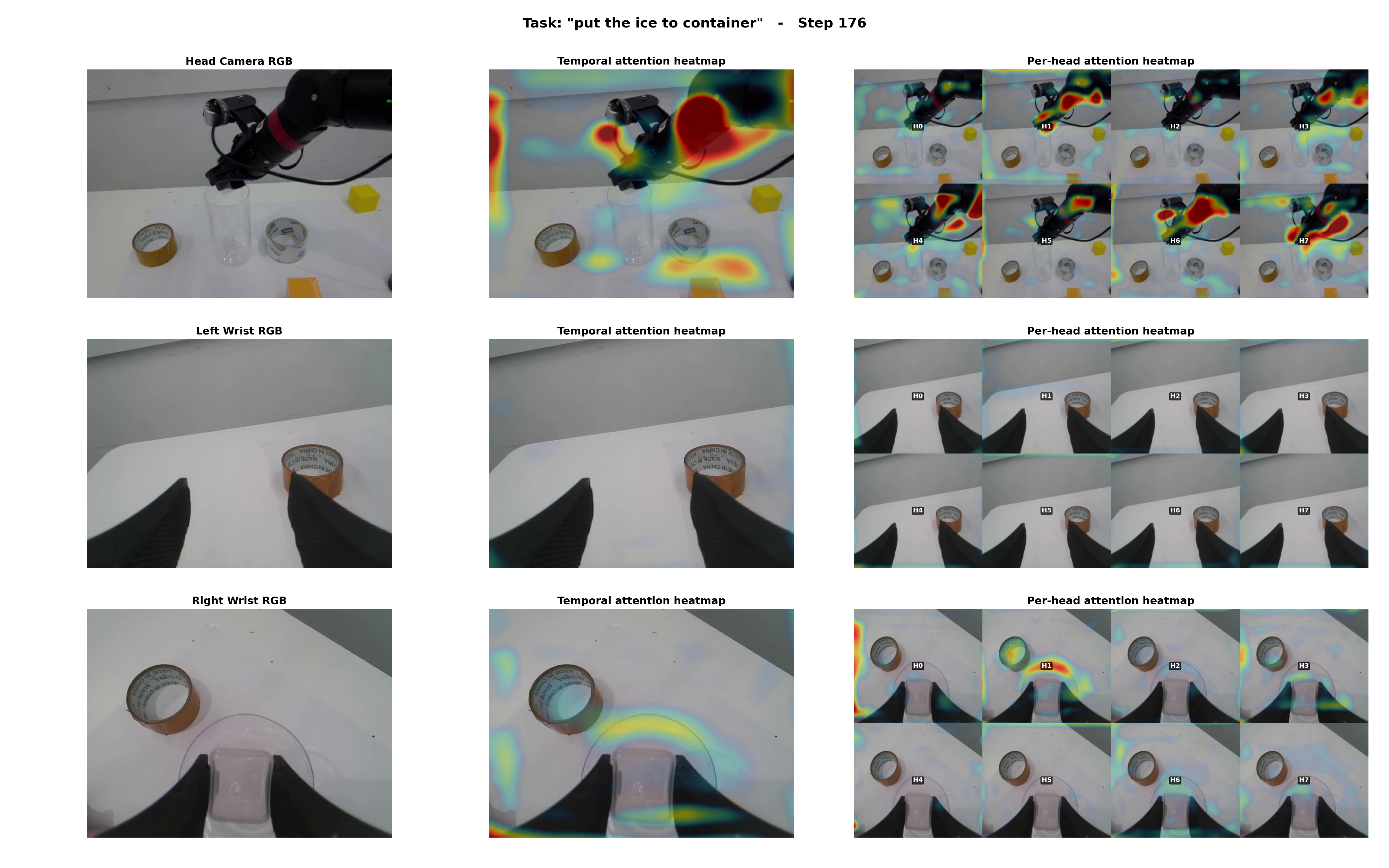}
    \caption{Multi-view attention visualization for a single-arm execution step in the task ``place the ice cube into the container.'' Because the left arm is inactive, the left-wrist view receives little high attention. In contrast, the right-wrist view concentrates attention around the container opening and active end-effector region, while the head view attends broadly to the robot arm and task objects.}
    \label{fig:single_arm_attention}
\end{figure}

\begin{table}[htbp]
\centering
\caption{End-to-end inference runtime. Measurements report mean $\pm$ standard deviation.}
\label{tab:runtime}
\begingroup
\scriptsize
\setlength{\tabcolsep}{3pt}
\resizebox{\linewidth}{!}{%
\begin{tabular}{lccccc}
\toprule
Method & Geo & Learned & State & H200 (ms / Hz) & RTX 4090 (ms / Hz) \\
\midrule
RGB-only backbone & \xmark & \xmark & \xmark & \res{71.9}$\pm$\res{3.1} / \res{13.9} & \res{109.2}$\pm$\res{1.8} / \res{9.2} \\
w/o state queries & \cmark & \cmark & \xmark & \res{91.4}$\pm$\res{4.5} / \res{10.9} & \res{145.2}$\pm$\res{7.0} / \res{6.9} \\
\textbf{GeoAlign} & \textbf{\cmark} & \xmark & \textbf{\cmark} & \res{92.1}$\pm$\res{3.8} / \res{10.9} & \res{138.9}$\pm$\res{1.2} / \res{7.2} \\
\bottomrule
\end{tabular}%
}
\par\smallskip
\begin{minipage}{0.88\linewidth}
\footnotesize
\emph{Notes.} Geo: geometry branch enabled. Learned and State indicate the query source used by the geometry branch.
\end{minipage}
\endgroup
\end{table}

The alignment module adds cross-attention between 8 queries and 5,476 spatial tokens per view, producing 8 extra conditioning tokens for the decoder. End-to-end runtime remains practical for deployment: on an H200 GPU, \method runs at \res{92.1}\,$\pm$\,\res{3.8} ms per action chunk (\res{10.9} Hz), compared with \res{71.9}\,$\pm$\,\res{3.1} ms (\res{13.9} Hz) for the RGB-only baseline.

\FloatBarrier
\section{Implementation Details}
\label{app:implementation}

\begin{algorithm}[htbp]
\caption{GeoAlign policy-training step after geometry post-training}
\label{alg:training}
\begin{algorithmic}[1]
\Require Demonstration batch $(I_t,\ell,s_t,A_t,m_t)$; frozen post-trained geometry encoder $E_{\mathrm{geo}}$; VLA encoder $E_{\mathrm{vlm}}$; state encoder $E_s$; alignment module; decoder $D_\theta$.
\State Encode semantic context $Z_t^{\mathrm{vlm}}\gets E_{\mathrm{vlm}}(I_t,\ell)$.
\State Extract \gep features $F_t^{\mathrm{geo}}\gets E_{\mathrm{geo}}(I_t)$; project them and add spatial/view embeddings to form $\Phi_t^{\mathrm{geo}}$.
\State Encode state $h_t\gets \LN(E_s(s_t))$ and generate query slots $Q_t\gets \reshape(W_qh_t,K,d_g)+P^q$.
\State Cross-attend $Q_t$ to $\Phi_t^{\mathrm{geo}}$ and project the outputs to compact geometry tokens $Z_t^{\mathrm{geo}}$.
\State Form decoder context $C_t\gets[Z_t^{\mathrm{vlm}};Z_t^{\mathrm{geo}}]$.
\State Sample $\epsilon\sim\mathcal{N}(0,I)$ and $\tau$ with the GR00T beta time sampler; set $x_\tau=(1-\tau)\epsilon+\tau A_t$.
\State Predict $\hat{v}_\theta\gets D_\theta(x_\tau,s_t,C_t,\tau)$.
\State Update trainable parameters using $\sum m_t\odot(\hat{v}_\theta-(A_t-\epsilon))^2/\sum m_t$.
\end{algorithmic}
\end{algorithm}

\paragraph{GEP feature branch details.}
The geometry branch uses a Depth Anything V2-Small model post-trained on robot RGB-D data. We refer to the DINOv2 visual backbone plus DA-V2 feature reassembly stage as the DA-V2 encoder-side feature extractor. Input images are resized to $518\times 518$ and patchified with patch size 14, yielding a $37\times 37$ base patch grid. After geometry post-training, the depth prediction head is discarded and \texttt{geometry\_feature\_stage}=1 is used as the geometry descriptor. This stage applies reassemble factor 2, producing a $74\times74$ feature map. A linear projection maps each reassembled feature to 256-dimensional geometry tokens, producing 5,476 spatial tokens per view, which are augmented with 2D sine-cosine positional embeddings. State-guided cross-attention uses 8 query tokens and 8 attention heads to produce 8 compact geometry tokens, each projected to 2,048 dimensions for the DiT action head.

\paragraph{Geometry-feature supervision protocol.}
The geometry branch is post-trained from Depth Anything V2-Small (24.8M parameters) on the RobbyVLA subset of LingBot-Depth~\citep{tan2026maskeddepth}. RobbyVLA contains 580,960 paired RGB-D frames collected during real-world VLA manipulation tasks with Franka and UR7e robot arms. The subset uses left and right RealSense D405 cameras and provides RGB images, raw sensor depth, ground-truth depth, and camera intrinsics. The geometry post-training data do not include LIBERO or SimplerEnv depth data, the evaluated ALOHA task trajectories, test object configurations, or deployment trials. The encoder is post-trained once and reused across the evaluated platforms. We use a 90/5/5 train/validation/test split (522,864 / 29,048 / 29,048 frames). Post-training fine-tunes the DA-V2 encoder-side feature extractor with a lower learning rate and the depth prediction layers with a higher learning rate, so that the model can predict metric depth in the robot workspace. For RGB image $I$, metric depth $D$, prediction $\hat{D}$, and $d_p=\log D_p-\log \hat{D}_p$, the valid-pixel depth objective is
\begin{equation}
    \mathcal{L}_{\mathrm{dep}}
    =
    \sqrt{
    \frac{1}{|\Omega|}\sum_{p\in\Omega} d_p^2
    -
    \lambda_{\mathrm{silog}}
    \left(\frac{1}{|\Omega|}\sum_{p\in\Omega} d_p\right)^2
    }.
\end{equation}
During VLA policy training, the post-trained geometry encoder is frozen and only the downstream projector, alignment module, and action head are trained with policy demonstrations. Table~\ref{tab:gep_training} reports the training configuration, and Table~\ref{tab:gep_metrics} reports held-out depth metrics.

\begin{table}[htbp]
\centering
\caption{Geometry post-training configuration.}
\label{tab:gep_training}
\begingroup
\small
\setlength{\tabcolsep}{4pt}
\begin{tabularx}{\linewidth}{L{0.31\linewidth}Y}
\toprule
Item & Value \\
\midrule
Base model & Depth Anything V2-Small with DA-V2 encoder-side feature extractor (24.8M parameters) \\
Training data & 580,960 paired RGB-D frames from the RobbyVLA subset of LingBot-Depth~\citep{tan2026maskeddepth} \\
Data source & Franka and UR7e VLA manipulation data with left/right RealSense D405 cameras \\
Train/validation/test split & 90/5/5: 522,864 / 29,048 / 29,048 frames \\
Epochs & 8 \\
Batch size & 32 \\
Learning rate & DA-V2 encoder-side feature extractor $5\times10^{-6}$; depth prediction layers $5\times10^{-5}$ \\
Optimizer & AdamW, weight decay 0.01, betas $(0.9,0.999)$ \\
Learning-rate schedule & cosine \\
Loss & SiLog loss, $\lambda=0.5$ \\
Image input & $518\times518$ lower-bound resize with aspect ratio preserved and crop \\
Depth range & 0.001--20.0 m \\
Augmentation & random horizontal flip, $p=0.5$ \\
Seed & 42 \\
\bottomrule
\end{tabularx}
\endgroup
\end{table}

\begin{table}[htbp]
\centering
\caption{Held-out depth validation for the geometry post-training stage on RobbyVLA.}
\label{tab:gep_metrics}
\begingroup
\small
\setlength{\tabcolsep}{4pt}
\resizebox{\linewidth}{!}{%
\begin{tabular}{lccccccccc}
\toprule
 & AbsRel & SqRel & RMSE & RMSE\_log & log10 & SiLog & $\delta_1$ & $\delta_2$ & $\delta_3$ \\
\midrule
Value & 0.1871 & 0.0684 & 0.1375 & 0.1741 & 0.0543 & 0.1561 & 0.8758 & 0.9352 & 0.9621 \\
\bottomrule
\end{tabular}%
}
\endgroup
\end{table}

Threshold accuracies $\delta_i$ use the standard $1.25^i$ criteria.

\paragraph{Policy training protocol.}
Table~\ref{tab:training_config} summarizes the ALOHA training and deployment configuration. Controlled GR00T variants share policy trajectories, action normalization, observation history, augmentation, optimizer, training-step budget, and checkpoint selection. RGB-only has no geometry branch. Geometry-feature variants freeze their encoder except for w/ unfrozen encoder; w/o post-training uses the unadapted DA-V2 encoder. RobbyVLA supervision is additional to the shared policy demonstrations.

\begin{table}[htbp]
\centering
\caption{ALOHA training and deployment configuration for GeoAlign and controlled baselines using Isaac-GR00T N1.6-3B~\citep{bjorck2025gr00tn1} and a DiT action head~\citep{peebles2023dit}. Camera views and state dimensions here describe ALOHA.}
\label{tab:training_config}
\begingroup
\small
\setlength{\tabcolsep}{4pt}
\begin{tabularx}{\linewidth}{L{0.31\linewidth}Y}
\toprule
Item & Value \\
\midrule
Model & Isaac-GR00T N1.6-3B with Eagle-Block2A-2B VLM, GEP geometry branch, and 32-layer DiT action head \\
Training steps & 20,000 \\
Global batch size & 640 \\
Learning rate & $1\times10^{-4}$ \\
Optimizer & AdamW, weight decay $1\times10^{-5}$ \\
Learning-rate schedule & cosine schedule with warmup ratio 0.05 \\
Precision / distributed training & BF16 mixed precision with DeepSpeed ZeRO-2 \\
Image preprocessing & resize short side to 256, center-crop to $244\times244$, resize to $224\times224$ \\
Camera inputs & 3 views: \texttt{cam\_high}, \texttt{cam\_left\_wrist}, \texttt{cam\_right\_wrist} \\
Action horizon & 16 steps, delta indices 0--15 \\
State dimension & 14: left arm 6 + left gripper 1 + right arm 6 + right gripper 1 \\
Action representation & relative joint deltas for arm joints; absolute gripper commands \\
Control frequency & 20 Hz \\
State dropout & 0.8 on encoded action-head state during training; geometry-query state is not dropped \\
Data augmentation & color jitter: brightness 0.3, contrast 0.4, saturation 0.5, hue 0.08 \\
Flow matching & 4-step Euler inference from Gaussian noise; Beta$(1.5,1.0)$ training flow-time sampler \\
Data seed & 42 \\
Hardware & 2$\times$ NVIDIA H200 GPUs \\
Training time & approximately 16 hours \\
\bottomrule
\end{tabularx}
\endgroup
\end{table}

\FloatBarrier
\section{Evaluation Protocol}
\label{app:protocol}

\paragraph{LIBERO protocol.}
We evaluate the official LIBERO Spatial, Object, Goal, and LIBERO-10 suites. Each suite contains 10 tasks, and each task is evaluated with 200 rollouts, for a total of 8,000 rollouts. We report suite-level episode success and the unweighted average over the four suites.

\paragraph{SimplerEnv-Fractal protocol.}
We evaluate three shared SimplerEnv Google Robot / Fractal task families: Pick Coke Can, Move Near, and Open/Close Drawer. Pick Coke Can and Move Near are evaluated with 200 rollouts each; Open/Close Drawer averages the Open Drawer and Close Drawer tasks, each evaluated with 200 rollouts using the task success criterion from the environment. We report task-family success and the unweighted average over the three task families.

\paragraph{Real-world scoring.}
We evaluate eight ALOHA tabletop tasks. The first three tasks use 40 collected training trajectories each, and the five plate-relative placement tasks use 10 collected training trajectories each. Each policy is deployed for 20 trials per task. Object and target poses are randomized within an approximately $20\text{ cm}\times20\text{ cm}$ reachable tabletop region around the nominal task pose, with yaw randomized when it does not invalidate the task. For the clear-tape task, success requires taking the tape from the partially covered box and holding it stably. For the transparent-bottle task, the ice cube must enter the bottle mouth and remain inside. For the tape-roll task, the small tape roll must be placed inside the inner opening of the big tape roll. For plate-relation tasks, the object center must end in the instructed relative region of the plate and remain stable after release.

\FloatBarrier
\section{Reproducibility Checklist}
\label{app:checklist}

The main text and appendices report the geometry post-training dataset size, train/validation/test splits, camera views, image resolution, action space, action horizon, base VLA checkpoint, geometry backbone, number of attention tokens, policy hyperparameters, random seed, hardware, SimplerEnv task list, ALOHA trial counts, and real-world success criteria. Tables~\ref{tab:gep_training}, \ref{tab:gep_metrics}, and~\ref{tab:training_config} summarize the geometry post-training, held-out depth validation, and policy training configurations.

\end{document}